\documentclass[10pt,twocolumn,letterpaper]{article}

\usepackage[pagenumbers]{cvpr} 

\usepackage{graphicx}
\usepackage{amsmath}
\usepackage{amssymb}
\usepackage{booktabs}
\usepackage{microtype}
\usepackage{algorithm}
\usepackage{algorithmic}
\usepackage{multirow}
\usepackage[dvipsnames]{xcolor}
\usepackage{makecell}
\usepackage{colortbl}
\definecolor{gray}{gray}{0.5}

\usepackage[pagebackref,breaklinks=true,colorlinks]{hyperref}

\usepackage[capitalize]{cleveref}
\crefname{section}{Sec.}{Secs.}
\Crefname{section}{Section}{Sections}
\Crefname{table}{Table}{Tables}
\crefname{table}{Tab.}{Tabs.}

\def\cvprPaperID{*****} 
\def\confName{CVPR}
\def\confYear{2022}

\begin{document}

\title{Just Leaf It: Accelerating Diffusion Classifiers with Hierarchical Class Pruning}

\author{Arundhati S. Shanbhag$^{2,3}$, Brian B. Moser$^{1, 2, 3}$, Tobias C. Nauen$^{1, 2}$, \\
Stanislav Frolov$^{1, 2}$, Federico Raue$^{1}$, and Andreas Dengel$^{1, 2}$\\
$^{1}$German Research Center for Artificial Intelligence \\
$^{2}$University of Kaiserslautern-Landau\\
$^{3}$Equal Contribution\\
{\tt\small first.last@dfki.de}
}
\maketitle

\begin{abstract}
Diffusion models, celebrated for their generative capabilities, have recently demonstrated surprising effectiveness in image classification tasks by using Bayes' theorem. 
Yet, current diffusion classifiers must evaluate every label candidate for each input, creating high computational costs that impede their use in large-scale applications.
To address this limitation, we propose a Hierarchical Diffusion Classifier (HDC) that exploits hierarchical label structures or well-defined parent–child relationships in the dataset.
By pruning irrelevant high-level categories and refining predictions only within relevant subcategories (leaf nodes and sub-trees), HDC reduces the total number of class evaluations.
As a result, HDC can speed up inference by as much as 60\% while preserving and sometimes even improving classification accuracy.
In summary, our work provides a tunable control mechanism between speed and precision, making diffusion-based classification more feasible for large-scale applications.
\end{abstract}


\section{Introduction}
\label{sec:intro}

Generative models in the visual domain aim to capture the entire data distribution, yielding a rich and detailed understanding of the underlying structure \cite{goodfellow2020generative, rezende2015variational}.
This broad understanding lets them create new content while revealing important data traits \cite{hinton2007recognize,moser2024latent}.
Within this landscape, diffusion models in particular stand out for producing high-fidelity images via an iterative Markov process of adding and removing noise \cite{moser2024diffusion, bar2023multidiffusion, frolov2024spotdiffusion, lugmayr2022repaint, ho2020denoising}. 

Yet, most generative methods over the past decade have focused on content generation rather than leveraging their discriminative potential \cite{frolov2021adversarial, lugmayr2022repaint, zhang2023adding, wu2023uncovering, betker2023improving}.
More recently, researchers have begun to repurpose pre-trained diffusion models for zero-shot classification, marking a shift toward using generative models as discriminators - known as diffusion classifiers - with interesting applications in training-free, open-set, and robust classification \cite{li2023your,chen2024your,clark2023, chen2024diffusion, allgeuer2024unconstrained}.
More specifically, diffusion models that learned $p \left( \mathbf{x} \mid \mathbf{c} \right)$ can be easily converted into classifiers by exploiting the Bayes' theorem to derive $p \left( \mathbf{c} \mid \mathbf{x} \right)$.
Thus, given an image $\mathbf{x}$ and a set of $N_C$ possible candidate classes $\{\mathbf{c}_i\}_{i=1}^{N_C}$, we can calculate the likelihood of $\mathbf{x}$ belonging to each class $\mathbf{c}_i$. 

\begin{figure}
    \begin{center}
        \includegraphics[width=\columnwidth]{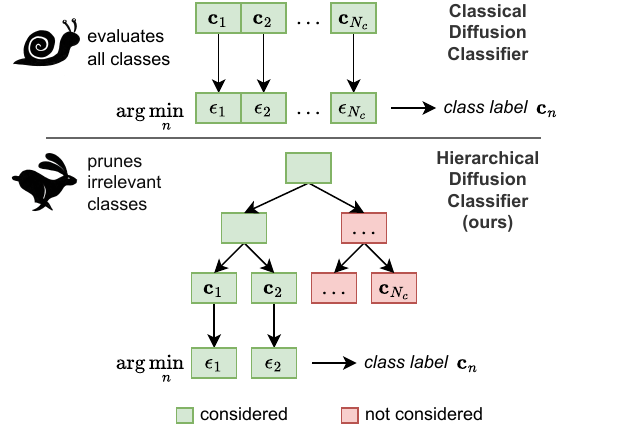}
        \caption{\label{fig:idea}
        Comparison between the classical diffusion classifier and our proposed Hierarchical Diffusion Classifier (HDC). 
        While the classical approach evaluates all possible classes to find the correct label, which leads to unnecessary computation, HDC prunes irrelevant classes early, focusing only on the most relevant candidates. 
        This hierarchical pruning reduces computational overhead and accelerates inference.
        }
    \end{center} 
\end{figure}

In practice, this means adding noise to $\mathbf{x}$ and estimating the expected loss of noise reconstruction via Monte Carlo, \textit{i.e.}, through repeated calculations and averaging.
This procedure, known as $\varepsilon$-prediction, has to be done for \textit{each} class label candidate.
Although Bayes' Theorem elegantly adapts diffusion models for zero-shot use, meaning they can classify without any additional training, the scaling with $N$ classes poses a considerable computational challenge \cite{ganguli2022predictability, clark2023, li2023your, moser2024latent}.

To alleviate the computational burden, we propose a training-free extension to diffusion classifiers that exploits a hierarchical search over label trees rather than evaluating each label individually, which we coined as Hierarchical Diffusion Classifier (HDC) and illustrated in \autoref{fig:idea}.
In the first stage, termed the pruning stage, HDC eliminates irrelevant branches by traversing the label tree level-by-level and keeping only the most promising synsets \textit{i.e.}, category-defining labels determined by the best $\varepsilon$-predictions.
In contrast to classical diffusion classification, the $\varepsilon$-predictions use fewer computation steps for the Monte Carlo estimate to save additional runtime.
Subsequently, HDC performs the classical diffusion classification on the remaining candidate leaf nodes.
As a result, HDC achieves a speed-up of roughly 60\%, saving hundreds of hours for ImageNet-1K \cite{deng2009imagenet} while maintaining similar accuracy. 
In addition, HDC can achieve better accuracy, \textit{i.e.}, 65.16\% per class instead of 64.90\%, by using roughly the same calculation time as traditional diffusion classifiers by using lower pruning ratios in the first stage. 
This efficiency gain makes HDC more practical for large-scale applications.

Overall, HDC introduces a new controllable balance between inference speed and classification accuracy by adjusting pruning factors, enabling diffusion classifiers to be flexibly used in large-scale discriminative tasks. 
By adopting a hierarchical approach, we provide a scalable and practical solution for utilizing diffusion models in a wide range of applications beyond their original generative purpose. HDC also potentially improves the interpretability of classification results in challenging domains.
The main contributions of our work can be summarized as follows:

\begin{enumerate}
    \item We demonstrate that the inference time of zero-shot diffusion classifiers can be significantly accelerated with a hierarchical label search, reducing computational complexity while maintaining similar accuracy. 
    \item We present a novel Hierarchical Diffusion Classifier (HDC) that leverages the label structure of datasets such as ImageNet-1K to narrow down candidate classes.
    \item Our HDC framework achieves faster inference times (roughly 60\%) while maintaining or improving classification performance compared to the classical diffusion method (\textit{i.e.}, 65.16\% instead of 64.90\%).
\end{enumerate}

\section{Related Work}
\label{sec:realted_work}
Diffusion models have disrupted the landscape of generative models, challenging the longstanding dominance of GANs \cite{goodfellow2020generative,frolov2021adversarial} and setting a new standard in generating high-quality, realistic data \cite{dhariwal2021diffusion}. 
Broadly speaking, they capture data distributions by iteratively adding and removing noise within a Markovian process.
Traditionally used to model data distributions, these generative models have recently been explored for their discriminative capabilities as well.

\textit{Li et al.} \cite{li2023your} introduced diffusion classifiers by using Stable Diffusion (SD) \cite{rombach2022high} as a zero-shot classifier without the need for additional training. 
SD, originally designed for text-to-image generation and trained on a subset of the LAION-5B dataset \cite{schuhmann2022}, leverages its ability to synthesize data to discriminate between images by evaluating prediction errors across class labels, \textit{i.e.}, $\varepsilon$-predictions. 
More specifically, they compute class scores based on differences in predicted and actual noise, offering an efficient classification method.
Similarly in spirit, \textit{Clark et al.} \cite{clark2023} further explored diffusion models like SD \cite{rombach2022high} and Imagen \cite{saharia2022photorealistic} for discriminative tasks by aggregating score matrices across class labels and timesteps. 
Using a weighted score function, they assign images to the class with the lowest aggregated score, demonstrating the transferability of generative representations to classification tasks.

While enabling the zero-shot diffusion classification, both approaches consider all classes as candidates to classify a single image, which is computationally inefficient.
Since inference time for zero-shot classification scales linearly with the number of classes, any computational improvement can significantly impact classifications on large-scale datasets like ImageNet-1K \cite{deng2009imagenet}. 
\textit{Li et al.} \cite{li2023your} address this by using a weak discriminative model to filter out obviously incorrect classes before performing zero-shot classification, thus speeding up the process. 
Similarly, \textit{Clark et al.} \cite{clark2023} employ a successive elimination strategy within a multi-armed bandit framework to iteratively narrow down the set of candidates.

Despite some reduction in computational complexity, they still process each class label at every diffusion timestep during inference. 
In contrast, our work explores the potential of leveraging hierarchical structures of datasets like ImageNet-1K. 
By integrating hierarchical pruning strategies, we progressively refine the set of candidate classes at each hierarchy level, allowing for faster and more accurate predictions.

\section{Methodology}
\label{sec:methodology}
\begin{figure*}
    \begin{center}
        \includegraphics[width=\textwidth]{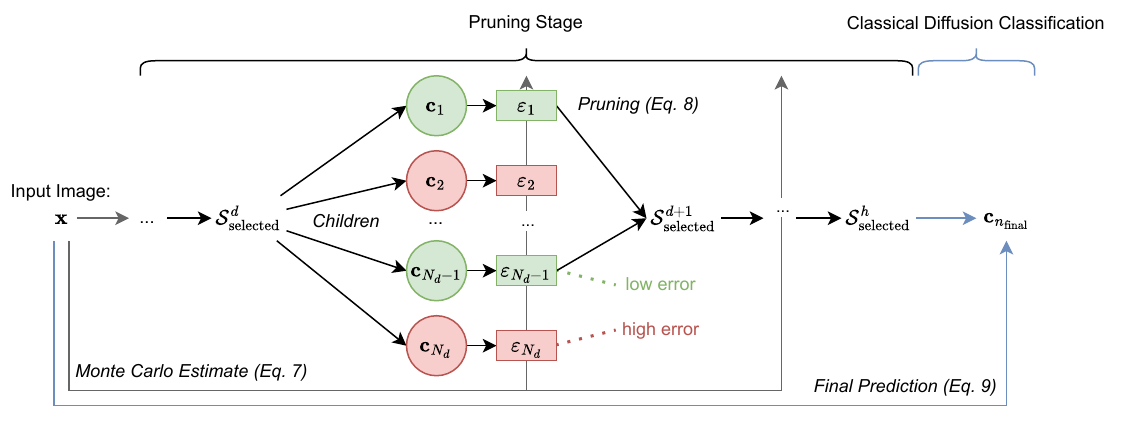}
        \caption{\label{fig:h_diff_classifier}
        Overview of our Hierarchical Diffusion Classifier (HDC). 
        Starting with an input image $\mathbf{x}$, noise $\varepsilon \sim \mathcal{N}(0, I)$ is added to generate a noisy image, resulting in $\mathbf{x}_t$ for multiple timesteps $t$. 
        Next, we use the diffusion classifier with a reduced number of $\varepsilon$-predictions and hierarchical conditioning prompts like ``A photo of a \{synclass / class name\}'' to progressively refine the classification through multiple levels of the label tree. By doing so, we keep track of the most promising classes (highlighted in green) and ignore the rest (highlighted in red).  The set of selected nodes during the pruning stage is denoted as $\mathcal{S}^d_{\text{selected}}$, where $d$ denotes the step count during traversal from 1 to $h$, the depth of the tree. Subsequently, the classical diffusion classifier pipeline is applied to the pruned, more specific subcategories (leaf nodes), which results in faster classification overall. 
        }
    \end{center} 
\end{figure*}

This section provides a brief overview of diffusion classifiers and introduces our proposed Hierarchical Diffusion Classifier (HDC), as shown in \autoref{fig:h_diff_classifier} and outlined in \autoref{alg:hierarchical_diff_classifier}. 

\subsection{Diffusion Classifier Preliminaries}
The diffusion classifier is based on the formulation introduced by \textit{Li et al.} \cite{li2023your}. 
The key idea is that, given a trained diffusion model $p_\theta$, we can leverage the predictions of the diffusion model, $p_\theta(\mathbf{x} \mid \mathbf{c}_i)$, to infer the probability of a class $\mathbf{c}_i$ given an input $\mathbf{x}$ using Bayes’ theorem to derive $p_\theta(\mathbf{c}_i \mid \mathbf{x})$. 
This can be expressed as:
\begin{align}
    p_\theta(\mathbf{c}_i \mid \mathbf{x}) = \frac{p(\mathbf{c}_i)\ p_\theta(\mathbf{x} \mid \mathbf{c}_i)}{\sum\limits^{N_C}_{j=1} p(\mathbf{c}_j)\ p_\theta(\mathbf{x} \mid \mathbf{c}_j)}
\label{eq:bayes}
\end{align}
Here, $p_\theta(\mathbf{x} \mid \mathbf{c}_i)$ is the likelihood of generating the input $\mathbf{x}$ given class $\mathbf{c}_i$, and $p(\mathbf{c}_i)$ is the prior probability of class $\mathbf{c}_i$.

To simplify this expression, we assume that the prior distribution over the classes is uniform, \textit{i.e.}, $p(\mathbf{c}_i) = \frac{1}{N_C}$ for $N_C$ classes. This assumption leads to the cancellation of the $p(\mathbf{c})$ terms, simplifying the expression in \autoref{eq:bayes} to:
\begin{align}
    p_\theta(\mathbf{c}_i \mid \mathbf{x}) = \frac{\ p_\theta(\mathbf{x} \mid \mathbf{c}_i)}{\sum\limits^{N_C}_{j=1} \ p_\theta(\mathbf{x} \mid \mathbf{c}_j)}
\label{eq:bayes_2}
\end{align}
Next, by exploiting the Evidence Lower Bound (ELBO), we can further refine \autoref{eq:bayes_2} into a more practical expression:
We approximate the likelihood $p_\theta(\mathbf{x} \mid \mathbf{c}_i)$ using the error between the noise $\varepsilon$ and the predicted noise $\varepsilon_\theta$ in the diffusion process. 
More specifically, we define
\begin{align}
    d \left( \varepsilon, \mathbf{x}, \mathbf{c} \right) = \|\varepsilon - \varepsilon_\theta(\mathbf{x}, \mathbf{c})\|^2
\end{align}
to calculate the distance between the error and the predicted error of denoising $\mathbf{x}$ under the class label $\mathbf{c}$.
This results in the following posterior distribution over $\{\mathbf{c}_i\}_{i=1}^{N_C}$:
\begin{align}
    p_\theta(\mathbf{c}_i \mid  \mathbf{x}) 
    &= \frac{\exp\{- \mathbb{E}_{t, \varepsilon}d \left( \varepsilon, \mathbf{x}_t, \mathbf{c}_i \right)\}}{\sum\limits^{N_C}_{j=1} \exp\{- \mathbb{E}_{t, \varepsilon}d \left( \varepsilon, \mathbf{x}_t, \mathbf{c}_j \right)\}} 
\label{eq:posterior}
\end{align}

However, there is still room for improvement. 
A key insight from \textit{Li et al.} is that for zero-shot classification, we are primarily interested in the relative differences between the prediction errors across different classes rather than the absolute error values for each class. 
This insight leads to a simplified version of the posterior \autoref{eq:posterior} as follows:
\begin{align}
\label{eq:paired} 
p_\theta(\mathbf{c}_i\mid\mathbf{x}) \approx \frac{1}{ \sum\limits^{N_C}_{j=1}\exp\left\{\mathbb{E}_{t, \varepsilon} \Delta\left(\varepsilon, \mathbf{x}_t, \mathbf{c}_i, \mathbf{c}_j\right) \right\}}, \\
\Delta\left(\varepsilon, \mathbf{x}_t, \mathbf{c}_i, \mathbf{c}_j\right) = d\left( \varepsilon, \mathbf{x}_t, \mathbf{c}_i \right) -d \left( \varepsilon, \mathbf{x}_t, \mathbf{c}_j \right) \nonumber
\end{align}
\noindent
This reduces the computational burden since classification is now based on error ranking.

Naturally, calculating the expectation value would lead to an unbiased Monte Carlo estimate. Specifically, we approximate the expectation $\mathbb{E}_{t, \varepsilon}$ by sampling $M$ pairs of $(t_i, \varepsilon_i)$, where $t_i$ is uniformly sampled from the range $[1, T]$ and $\varepsilon_i$ is drawn from a standard normal distribution, $\varepsilon_i \sim \mathcal{N}(0, I)$. Using these samples, we make the following approximation:
\begin{align}
\label{eq:monte_carlo} 
    \mathbb{E}_{t, \varepsilon}d \left( \varepsilon, \mathbf{x}_t, \mathbf{c}_j \right)
    \approx \frac{1}{M}\sum_{i=1}^M d \left( \varepsilon_i, \widetilde{\mathbf{x}}_i, \mathbf{c} \right), \\
    \widetilde{\mathbf{x}}_i = \sqrt{\bar \alpha_{t_i}}\mathbf{x} + \sqrt{1-\bar\alpha_{t_i}} \varepsilon_i \nonumber
\end{align}

Moreover, instead of using different random samples of $(t_i, \varepsilon_i)$ to compute the ELBO for each conditioning input $\mathbf{c}_i$, we can also take advantage of a fixed set of samples $S = {(t_i, \varepsilon_i)}_{i=1}^M$. 
As a result, the error estimation is now consistent across all class conditions.
By plugging \autoref{eq:monte_carlo} into \autoref{eq:paired}, we can extract a diffusion classifier from any conditional diffusion model, such as Stable Diffusion \cite{rombach2022high}. 
This extracted diffusion classifier operates in a zero-shot manner, meaning it can classify without additional training on labeled data, even open-set vocabulary \cite{allgeuer2024unconstrained}.

\subsection{Hierarchical Diffusion Classifier (HDC)}
As previously shown, traditional diffusion classifiers need to evaluate all possible classes, which can be computationally expensive and time-consuming.
To ease the computational burden, we propose a Hierarchical Diffusion Classifier (HDC), which leverages the hierarchical label structure of a dataset to perform more efficient and accurate classification.

The core idea is to evaluate labels hierarchically and to progressively narrow down the possible candidate classes by pruning higher-level categories (such as ``animals'' or ``tools'') into more specific categories and classes (such as ``Hammerhead Shark'' or ``Screwdriver''). 
The higher-level categories are called ``synonym-sets'' or ``synsets''.
By iterating over the labels hierarchically, we can significantly reduce the number of classes that need to be evaluated, leading to faster predictions with potentially higher accuracy. 

More formally, let $T_h = \left( N, E \right)$ represent a hierarchical label tree of depth $h$, nodes $N$, and edges $E$.
Each node $n \in N$ in the tree corresponds to a synset (or class for leaf nodes), and $n_{\text{root}}$ is the root.
Moreover, let $\texttt{Children} \left(n \right) \subseteq N$ denote the set of child nodes of $n$, and $\mathbf{c}_n$ represent the synset or class label of a node $n$.
We set $\texttt{Children} \left(n \right) := n$ if $n$ is a leaf node to address imbalanced label trees.

Our proposed HDC aims to prune irrelevant classes and only considers more relevant classes (nodes) as we descend the tree within a selected set of nodes.
The set of selected nodes is denoted as $\mathcal{S}^d_{\text{selected}}$, where $d$ denotes the traverse step count starting from 1 and ending in $h$ (depth of the label tree). 
We start with the root node $n_{\text{root}}$, \textit{i.e.}, $\mathcal{S}^1_{\text{selected}} = \left\{ n_{\text{root}} \right\}$, which contains the highest-level categories as children. 

For each traverse step $d$, we evaluate recursively the error score for each child node of the selected nodes $n_s \in \mathcal{S}^d_{\text{selected}}$:
\begin{align}
\label{eq:childerrors}
&\forall n_s \in \mathcal{S}^d_{\text{selected}} : \, \forall n \in \texttt{Children} \left( n_s\right): \\ 
&\epsilon_n = \mathbb{E}_{t, \varepsilon} d \left( \varepsilon, \mathbf{x}_t, \mathbf{c}_n \right). \nonumber 
\end{align}
We use again Monte Carlo, \textit{i.e.}, \autoref{eq:monte_carlo}, to calculate $\epsilon_n$, but employ a smaller number of samples $M$ than in the classical diffusion classifier.
Instead of selecting a single node, we proceed with a set of nodes with the lowest error scores. 
This set of selected nodes is determined by a pruning strategy, where only the most relevant nodes are kept.

Formally, the set of the next selected nodes $\mathcal{S}^{d+1}_{\text{selected}}$ at each stage $d$ is defined as
\begin{align}
\label{eq:nextselected}
\mathcal{S}^{d+1}_{\text{selected}} = \{ n \in \texttt{Children} \left( n_s \right) & \mid  n_s \in \mathcal{S}^d_{\text{selected}}, \\
& \land \epsilon_n \leq \text{threshold}(K_d) \}. \nonumber
\end{align}
The pruning ratio $K_d$ determines the threshold, which dictates how many nodes from the current set are kept for the next level $d+1$ of the hierarchy.
Essentially, we use the threshold to act as a top-k pruning.
The pruning procedure is outlined in \autoref{alg:hierarchical_diff_classifier}.

\begin{algorithm}[t]
\caption{Hierarchical Diffusion Classifier (HDC) in the pruning stage for classifying one image}
\label{alg:hierarchical_diff_classifier}
\textbf{Input:} test image $\mathbf{x}$, $T_h = \left( N, E \right)$ with nodes $N$, edges $E$ and depth $h$, root node $n_{\text{root}}$, label inputs $\{\mathbf{c}_i\}_{i=1}^{N_c}$, pruning ratios $K_d$, and number of random samples $M$ (see \autoref{eq:monte_carlo}).
\begin{algorithmic}[1]
    \STATE \textcolor{gray}{\textit{// initialization}}    
    \STATE \texttt{Selected} = \texttt{list}(\texttt{Children}($n_{\text{root}}$))
    \STATE \texttt{Errors} = \texttt{dict}()
    \STATE \texttt{ErrorsCalculated} = \texttt{dict}()
    \FOR{ each node $n \in N$}
        \STATE \texttt{Errors}[$\mathbf{c}_n$] = \texttt{list}() 
        \STATE \texttt{ErrorsCalculated}[$\mathbf{c}_n$] = false
    \ENDFOR
    \STATE
    \STATE \textcolor{gray}{\textit{// modified diffusion classifier error calculations}} 
    \FOR{tree depth $d = 1, \dots, h$}
        \FOR{stage $i = 1, \dots, M$}
            \STATE Sample $t \sim [1, 1000]$
            \STATE Sample $\varepsilon \sim \mathcal{N}(0, I)$
            \STATE $\mathbf{x}_t = \sqrt{\bar{\alpha}_{t}} \mathbf{x} + \sqrt{1-\bar{\alpha}_{t}} \varepsilon$
            \STATE
            \STATE \textcolor{gray}{\textit{// calculate child errors (\autoref{eq:childerrors})}}
            \FOR{each node $n_s$ in \texttt{Selected}}
                \FOR{each child node $n \in $ \texttt{Children}($n_s$)}
                    \STATE \textcolor{gray}{\textit{// check if error already calculated}} 
                    \IF {\texttt{ErrorsCalculated[$\mathbf{c}_n$]}}
                        \STATE continue
                    \ENDIF
                    \STATE 
                    \STATE \texttt{Errors}[$\mathbf{c}_n$].\texttt{append}($\|\varepsilon - \varepsilon_\theta(\mathbf{x}_t, \mathbf{c}_n)\|^2$)
                \ENDFOR
            \ENDFOR
        \ENDFOR
        \STATE
        \STATE \textcolor{gray}{\textit{// descend in the tree and select top-k (\autoref{eq:nextselected})}}
        \STATE \texttt{ErrorsCalculated[\texttt{Selected}]} = true
        \STATE \texttt{SelErrors} = \texttt{mean} (\texttt{Errors[\texttt{Selected}]})
        \STATE \texttt{Selected} = $\texttt{TopK} \left( \texttt{SelErrors}, K=K_d \right)$
    \ENDFOR
    \STATE
    \STATE \textcolor{gray}{\textit{// return pruned class label set}}
    \STATE \textbf{Return:} \texttt{Selected}
\end{algorithmic}
\end{algorithm}

Eventually, this process reaches the leaf nodes at $d= h$, corresponding to actual class labels. 
At this point, we derived a pruned class set, which will then subsequently be used in the classical diffusion classifier pipeline with the original number of samples $M$ to determine the final class label:
\begin{align}
\label{eq:final}
\mathbf{c}_{n_\text{final}}, \quad \text{where } n_\text{final} = \arg\min_{n \in \mathcal{S}^{h}_{\text{selected}}} \epsilon_{n}.
\end{align}

\begin{figure}[t!]
    \begin{center}
        \includegraphics[width=\columnwidth]{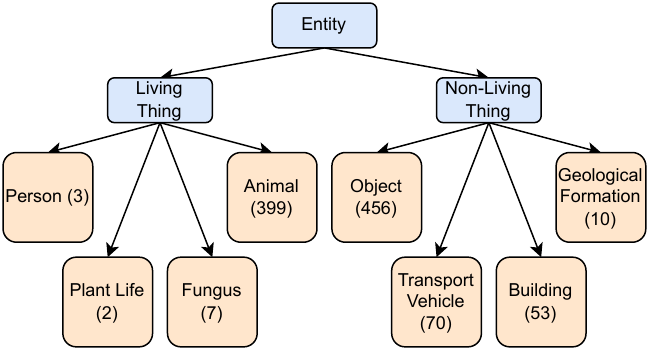}
        \caption{\label{fig:imagenet_hierarchy}
        Visualization of the ImageNet1K hierarchy, illustrating the first three levels of its tree structure. The categories are organized from broad entities (\textit{e.g.}, living and non-living things) to more specific groups (\textit{e.g.}, animals, objects, and transport vehicles), with the numbers in parentheses representing the total number of actual classes within each group.
        }
    \end{center} 
\end{figure}

\subsection{Tree Setup}
Our proposed HDC leverages the WordNet hierarchy upon which the ImageNet-1K ontology is constructed \cite{deng2009imagenet}. 
The images in the ImageNet-1K dataset are grouped into ``synonym-sets'' or ``synsets'' with 12 subtrees comprising around 80,000 synsets. 
To begin with, we created a hierarchical prompt list using the hierarchy from \textit{Engstrom et al.} \cite{robustness}. 
However, our proposed HDC approach is versatile and can utilize parent-child relationships even in the absence of explicit hierarchies within a dataset (\textit{e.g.}, using large language models). 
To demonstrate this empirically, we also constructed a hierarchy for the CIFAR-100 \cite{krizhevsky2009learning} dataset.

In general, the inference time of HDC increases with the depth of the tree. This is also observed in other tree-based classifiers \cite{10562290, 598994,6909245}. 
To optimize the trade-off between model performance and inference speed, we limit the maximum depth of the tree. 
We further modify the existing WordNet tree to better suit the classification objective by choosing sysnet labels with more definite meanings. Our final ImageNet-1K hierarchy tree has a depth of 7 levels. 

In the pruning stage of HDC, we iterate over the ImageNet-1K tree starting from nodes at level 3 of the hierarchy (``entity'' $\to$ \{``living-thing'', ``non-living thing''\} $\to$ \{``animals'', $\dots$\}), as shown in \autoref{fig:imagenet_hierarchy}. 
Starting at level 2 (``living thing'' vs.\ ``non-living thing'') showed no variation in error scores but increased inference time across all experiments indicating that well-defined sysnet labels perform better than vague ones. 

\subsection{Pruning Strategies}
\label{sec:pssetup}
Our proposed HDC method allows many pruning strategies to be implemented that balance accuracy and computational efficiency. 
We implemented two primary pruning strategies, one that works with fixed ratios of pruned nodes and one that adapts dynamically depending on the distribution of error predictions. 
In more detail:

\begin{itemize}
    \item \textbf{Strategy 1 - Fixed Pruning:} We select the top-k nodes with the lowest errors at each hierarchy level, defined by a pruning ratio $K_d$. 
    \item \textbf{Strategy 2 - Dynamic Pruning:} We keep only nodes within two standard deviations of the minimum error at each level, allowing a more adaptive, data-driven selection. 
\end{itemize}
\noindent
In contrast to Dynamic Pruning, Fixed Pruning allows for finer control over the trade-off between accuracy and runtime.
Our proposed pruning strategies offer varying degrees of control over the balance between accuracy and runtime, adapting to unique hierarchical structures for greater scalability.
Unlike traditional diffusion classifiers, which evaluate all classes for each input image, our pruning strategies strategically select candidate classes at each level, reducing computational load while maintaining similar class precision.

\subsection{Modification of Classes}
HDC also offers advantages in handling open-set scenarios and dynamic class modifications (\textit{i.e.}, adding and removing class labels without additional training), making it both flexible and efficient \cite{allgeuer2024unconstrained}. 
Since class labels are always represented as leaf nodes, removing a label is a straightforward process. Additionally, new labels can be seamlessly integrated into the tree either by attaching them as direct child nodes under the root or by leveraging the HDC itself to insert them in an optimal position using a greedy strategy. This approach ensures scalability, allowing the hierarchy to evolve dynamically while maintaining its structure and efficiency.

\section{Experimental Setup}
\label{sec:experimental_setup}
This section describes our experimental setup for testing the performance and reliability of HDC on ImageNet-1K and CIFAR-100, which includes a discussion about how to construct the hierarchical label tree and specifics to the classifier itself, prompting, and pruning strategies. 
Our code can be found on GitHub: \url{https://github.com/arus23/hierarchical_diffusion_classifier}.

\begin{table*}[t!]
\centering
\begin{tabular}{lccccc}
\toprule
{Method} & {Top 1 [\%]} & {Top 3 [\%]} & {Top 5 [\%]} & {Time [s]} & {Speed-Up [\%]} \\ 
\midrule
Diffusion Classifier (baseline) \cite{li2023your} & \underline{64.70} & \textbf{84.30} & \textbf{89.70} & 1600 & \textcolor{ForestGreen}{-} \\
HDC Strategy 1 (ours) & \textbf{64.90} & \underline{81.80} & \underline{86.30} & \underline{980}  & \textcolor{ForestGreen}{\underline{38.75}} \\ 
HDC Strategy 2 (ours) & 63.20 & 82.30 & 86.30 & \textbf{650} & \textcolor{ForestGreen}{\textbf{59.38}} \\ 
\bottomrule
\end{tabular}
\caption{Comparison of classification accuracy and inference time between the classical diffusion classifier \cite{li2023your} and our proposed HDC using two pruning strategies with Stable Diffusion 2.0 for \textbf{ImageNet-1K}. Both strategies demonstrate that HDC can significantly reduce classification time, achieving up to a 60\% speed-up in inference time with minimal impact on accuracy or reach even better top-1 precision with minimal impact on runtime.
The best results are marked in bold, second-best underlined. }
\label{tab:table1}
\end{table*}

\subsection{Classifier Setup}
HDC is based on the efficient framework established by \textit{Li et al.} \cite{li2023your}, with added modifications tailored for hierarchical processing and pruning of candidate classes, further customized for diffusion classification on Stable Diffusion (SD) \cite{rombach2022high}. 
Yet, our method is adaptable, allowing seamless integration with different diffusion models and possible fine-tuning to support various hierarchical pruning strategies.
To demonstrate this, we accommodate the SD versions 1.4, 2.0, and 2.1.
For Strategy 1 in our pruning setup, we set $K_d=0.5$ for all possible $d$-values. 
All evaluations were performed at 512$\times$512, the resolution under which all versions of SD were originally trained. 
Also following \textit{Li et al.}, we used the $l_2$ norm to compute the $\varepsilon_t$-predictions and sampled the timesteps uniformly sampled from $[1, 1000]$.

\begin{table*}[!t]
\centering
\begin{tabular}{lcccccccc}
\toprule
\multirow{3}{*}{{SD Version}} & \multicolumn{4}{c}{{Strategy 1}} & \multicolumn{4}{c}{{Strategy 2}} \\
\cmidrule(lr){2-5} \cmidrule(l){6-9}
 & {Top 1 [\%]} & {Top 1 [\%]} & {Time [s]}& {Speed-Up [\%]} & {Top 1 [\%]} & {Top 1 [\%]} & {Time [s]}& {Speed-Up [\%]} \\ 
 & (class-wise) & (overall) & & &  (class-wise) & (overall) & & \\
\midrule
SD 1.4 & 52.71 & 52.60 & 1000 & \textcolor{ForestGreen}{37.50} & 54.77 & 54.80 & \textbf{710} & \textcolor{ForestGreen}{\textbf{55.63}} \\ 
SD 2.0 & \textbf{65.16}  & \textbf{64.90} & 980 & \textcolor{ForestGreen}{38.75} & \textbf{63.33}  & \textbf{63.20} & 980 & \textcolor{ForestGreen}{38.75} \\ 
SD 2.1 & 61.15 & 61.00 & \textbf{950} & \textcolor{ForestGreen}{\textbf{40.63}} & 60.91 & 60.70 & 720 & \textcolor{ForestGreen}{55.00} \\ 
\bottomrule
\end{tabular}
\caption{Performance comparison of the HDC \textbf{with different SD versions} using Strategy 1 and Strategy 2 for \textbf{ImageNet-1K}. Top-1 accuracy and inference time (in seconds) are reported for each SD version, highlighting SD 2.0 as achieving the highest accuracy, while Strategy 2 in SD 1.4 yields the fastest inference time.}
\label{tab:sd_table}
\end{table*}

\begin{table}[!h]
\centering
\resizebox{\columnwidth}{!}{%
\begin{tabular}{lccccc}
    \toprule
    & Baseline & Strategy 1 & \multicolumn{3}{c}{Strategy 2 (Ours)} \\
    \cmidrule(lr){2-6}
    & (DC) & (Ours) & $K_d=0.75$ & $K_d=0.5$ & $K_d=0.4$ \\
    \midrule
    Class-Acc [\%] & \underline{68.93} & 65.12 & 56.79 & 60.57 & \textbf{72.23} \\
    Runtime [s/class] & 1000 & 740 & \textbf{275} & \underline{550} & 660 \\
    Speed-Up [\%] & \textcolor{ForestGreen}{-} & \textcolor{ForestGreen}{26.00} & \textcolor{ForestGreen}{\textbf{72.50}} & \textcolor{ForestGreen}{\underline{45.00}} & \textcolor{ForestGreen}{34.00} \\
    
    \bottomrule
\end{tabular}
}
\caption{Performance Comparison of the HDC on \textbf{CIFAR-100 with a self-generated label-tree} for Strategy 1 and Strategy 2 using SD 2.0. Our HDC shows improved runtime while improving accuracy (+3.3 percentage points and ca. 34\% speed-up with Strategy 2; $K_d=0.4$).}
\label{tab:cifar-table}
\end{table}

\subsection{Prompt Engineering}
The class labels are converted to the form ``a photo of a $<$\textit{class label}$>$'' using the template from the original work \cite{li2023your}. 
Inspired by \textit{Radford et al.} \cite{radford2021}, we also experiment with prompt templates ``A bad photo of a $<$\textit{class label}$>$'', ``A low-resolution photo of a $<$\textit{class label}$>$'' and ``itap of a $<$\textit{class label}$>$'' for ImageNet-1K. 
For CIFAR-100, we use ``a blurry photo of $<$\textit{class label}$>$''.

\section{Results}
This section presents our experimental results, evaluating different aspects of HDC, which were outlined previously: pruning strategies, prompt engineering, SD variations, and, finally, an overall evaluation of per-class accuracy.

\subsection{Pruning Strategies}
\autoref{tab:table1} highlights the results of our HDC across different pruning strategies compared to the classical diffusion classifier. 
As observed, both pruning strategies show significant improvements in runtime compared to classical diffusion classifiers, and each is suited to different prioritizations of speed versus accuracy.

Strategy 1 yields the best trade-off results on ImageNet-1K, achieving significant runtime reductions (up to 980 seconds) with a top-1 accuracy boost of 0.20 percentage points. 
By employing Strategy 2 (selecting candidates based on two standard deviations from the lowest error), we reduce the inference time even further to 650 seconds, though at the cost of a slight accuracy drop (\textit{i.e.}, 1.50 percentage points). 
Strategy 2 demonstrates that faster inference can be achieved with a small compromise in precision. 
Correspondingly, similar patterns are observed in \autoref{tab:cifar-table} for CIFAR-100. 

\begin{figure}[t!]
    \begin{center}
        \includegraphics[width=.95\columnwidth]{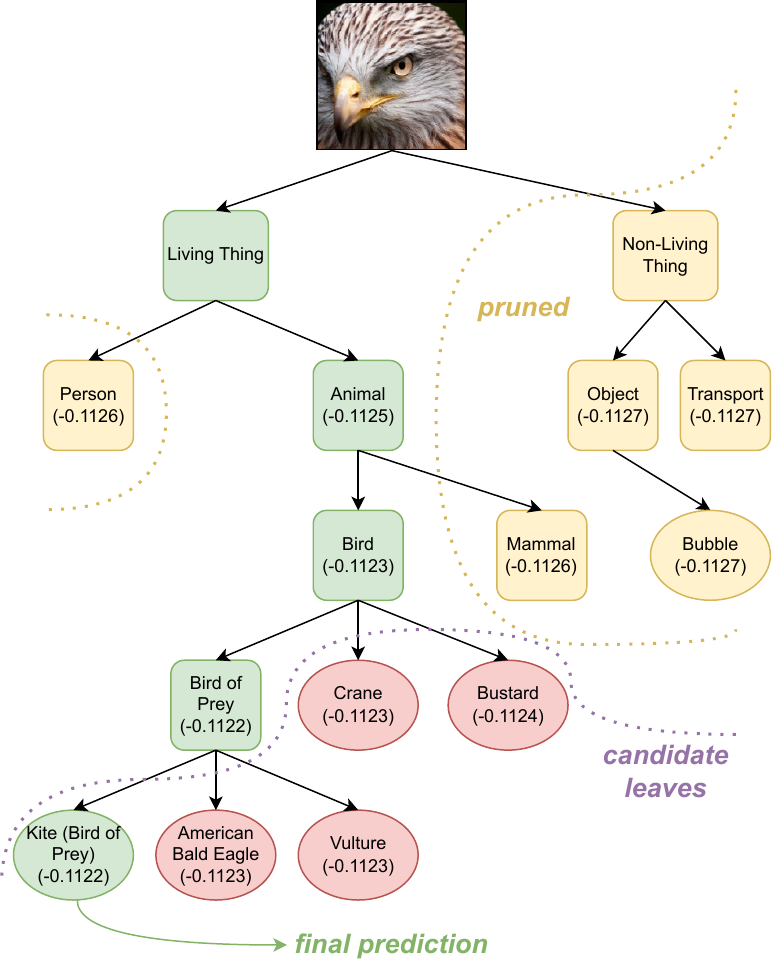}
        \caption{\label{fig:HDC_example}
        \textbf{Example classification} of an image in the pruning stage of the HDC using Strategy 1. 
        In this stage, the error scores of each node are used to iteratively prune the tree, narrowing it down to relevant leaf nodes that will undergo further refinement in subsequent stages. 
        The subsequent steps then focus on closely related nodes (see leaves under the purple line), such as the American Bald Eagle and Vulture, ultimately selecting the leaf node with the lowest error score—Kite (Bird of Prey) - in the final stage. 
        }
    \end{center} 
\end{figure}

In \autoref{fig:HDC_example}, we provide an empirical example of how HDC traverses the label tree in the pruning stage and calculates the error of candidate classes for final prediction (\textit{i.e.}, classical diffusion classification on pruned leaf nodes).

\subsection{Stable Diffusion Versions}
We evaluated the HDC using different SD versions to assess its flexibility and performance across generative backbones, as summarized in \autoref{tab:sd_table}. 
The results reveal that SD 2.0 provides the best trade-off between accuracy and inference time. 
Specifically, when using Strategy 1, SD 2.0 achieved the highest Top-1 accuracy at 64.14\% with an inference time of 980 seconds. 
In contrast, SD 1.4 demonstrates the fastest inference time of 710 seconds when paired with Strategy 2, albeit with a significant top-1 class-accuracy reduction to 54.77\%.

\begin{table}[t!]
\centering
\resizebox{\columnwidth}{!}{%
\begin{tabular}{clccc}
\toprule
{Strategy} & {Prompt-Type} & {Top 1 [\%]} & {Top 3 [\%]} & {Top 5 [\%]}  \\ 
\cmidrule(r){1-2} \cmidrule(l){3-5}
\multirow{4}{*}{{1}} & ``A photo of a $<$\textit{class label}$>$'' & \textbf{64.90} & \textbf{80.20} & \textbf{85.30} \\ 
&``A bad photo of a $<$\textit{class label}$>$''  & 59.90 & 79.60 & 84.90 \\
&``itap of a $<$\textit{class label}$>$'' & 61.37 & 81.33 & 86.30 \\ 
&``A low-resolution photo of a $<$\textit{class label}$>$'' & 57.50 & 76.46 & 80.94 \\
\cmidrule(r){1-5}
\multirow{4}{*}{{2}} &``A photo of a $<$\textit{class label}$>$'' & \textbf{63.20} & \textbf{82.30} & \textbf{86.30} \\ 
&``A bad photo of a $<$\textit{class label}$>$''  & 62.30 & 80.10 & 85.90 \\
&``itap of a $<$\textit{class label}$>$'' & 57.80 & 78.20 & 82.30 \\ 
&``A low-resolution photo of a $<$\textit{class label}$>$'' & 57.50 & 76.46 & 80.94 \\ 
\bottomrule
\end{tabular}
}
\caption{Evaluation \textbf{across different prompt types} for HDC using pruning Strategies 1 and 2 on \textbf{ImageNet-1K}. The standard prompt, ``A photo of a $<$\textit{class label}$>$'', consistently yields the highest Top-1, Top-3, and Top-5 accuracy. Alternative prompts, such as ``A bad photo of a $<$\textit{class label}$>$'' and ``A low-resolution photo of a $<$\textit{class label}$>$'', result in slight decreases in accuracy, showing that prompt variations can impact model performance. }
\label{tab:prompt_table}
\end{table}

\subsection{Prompt Engineering}
Inspired by \textit{Radford et al.} \cite{radford2021}, we also evaluated different prompt templates to assess their impact on accuracy and inference time, as shown in \autoref{tab:prompt_table}. 
The default prompt, “a photo of a $<$\textit{class label}$>$,” consistently achieved the best performance, suggesting that a straightforward prompt yields robust results across classes. Other templates, such as “a bad photo of a $<$\textit{class label}$>$” and “a low-resolution photo of a $<$\textit{class label}$>$,” resulted in a slight drop in accuracy without significantly affecting inference time.

The rationale for testing alternative prompts stems from a hypothesis that prompts hinting at lower-quality images might help the classifier generalize better to real-world cases with variable quality, capturing diverse visual characteristics. 
For instance, using terms like “bad” or “low-resolution” was expected to enhance robustness to noisy or degraded inputs.

Interestingly, however, the results show that the simpler, unmodified prompt performs best, indicating that the hierarchical model likely benefits from a more neutral prompt format when dealing with high-quality image data like ImageNet-1K. 
Nevertheless, these prompt variations may still hold potential for datasets with inherently low-resolution or distorted images, where quality-based prompts could help the classifier learn more generalized features.

We also observed a significant disparity in inference times across specific classes, such as “snail” (221 seconds) versus “keyboard space bar” (1400 seconds). 
This difference likely reflects the complexity of visual features within each category: classes with intricate or ambiguous features may require longer processing times due to the hierarchical classification structure. 

\begin{table}[!t]
\centering
\resizebox{\columnwidth}{!}{%
\begin{tabular}{lcc c}
\toprule
{Method} & {Avg. Accuracy [\%]} & {Time [s]}& {Speed-Up [\%]} \\ 
\cmidrule(r){1-1} \cmidrule(l){2-4}
Diffusion Classifier & 64.90 & 1600 & \textcolor{ForestGreen}{-}  \\ 
HDC Strategy 1 (ours)& \textbf{65.16} & 980 & \textcolor{ForestGreen}{38.75} \\ 
HDC Strategy 2 (ours)& 63.33 & \textbf{650} & \textcolor{ForestGreen}{\textbf{59.38}} \\ 
\bottomrule
\end{tabular}
}
\caption{Comparison of the average classification accuracy and inference time \textbf{per class on ImageNet-1K} for the classical diffusion classifier and HDC with Strategies 1 and 2 using SD 2.0. Strategy 1 achieves the highest per-class accuracy at K = 0.5, while Strategy 2 offers the fastest inference time with minimal accuracy loss.}
\label{tab:table2}
\end{table}

\subsection{Overall Accuracy vs. Inference Time}
In summary, \autoref{tab:table2} shows the overall accuracy and inference time across different pruning strategies. 
The baseline diffusion classifier achieves an accuracy of 64.90\% with an inference time of 1600 seconds, providing a reference for both speed and precision.

Using Strategy 1 in HDC demonstrates new state-of-the-art accuracy for diffusion classifiers with 65.16\%, while reducing the inference time by nearly 40\% to 980 seconds. 
This indicates that HDC can not only improve classification performance but also leads to a considerable reduction in computation. 
Reducing processing time while maintaining similar accuracy makes Strategy 1 a balanced choice for high-accuracy applications where inference speed is a priority.

Similarly, HDC with Strategy 2 leverages dynamic pruning to further accelerate inference. 
While it records a slight drop in accuracy to 63.33\%, Strategy 2 reduces inference time to 650 seconds - approximately 60\% faster than the baseline. 
This strategy demonstrates the potential of HDC for use cases requiring faster response times, with only a marginal trade-off in classification performance.

\begin{figure}[t!]
    \begin{center}
        \includegraphics[width=\columnwidth]{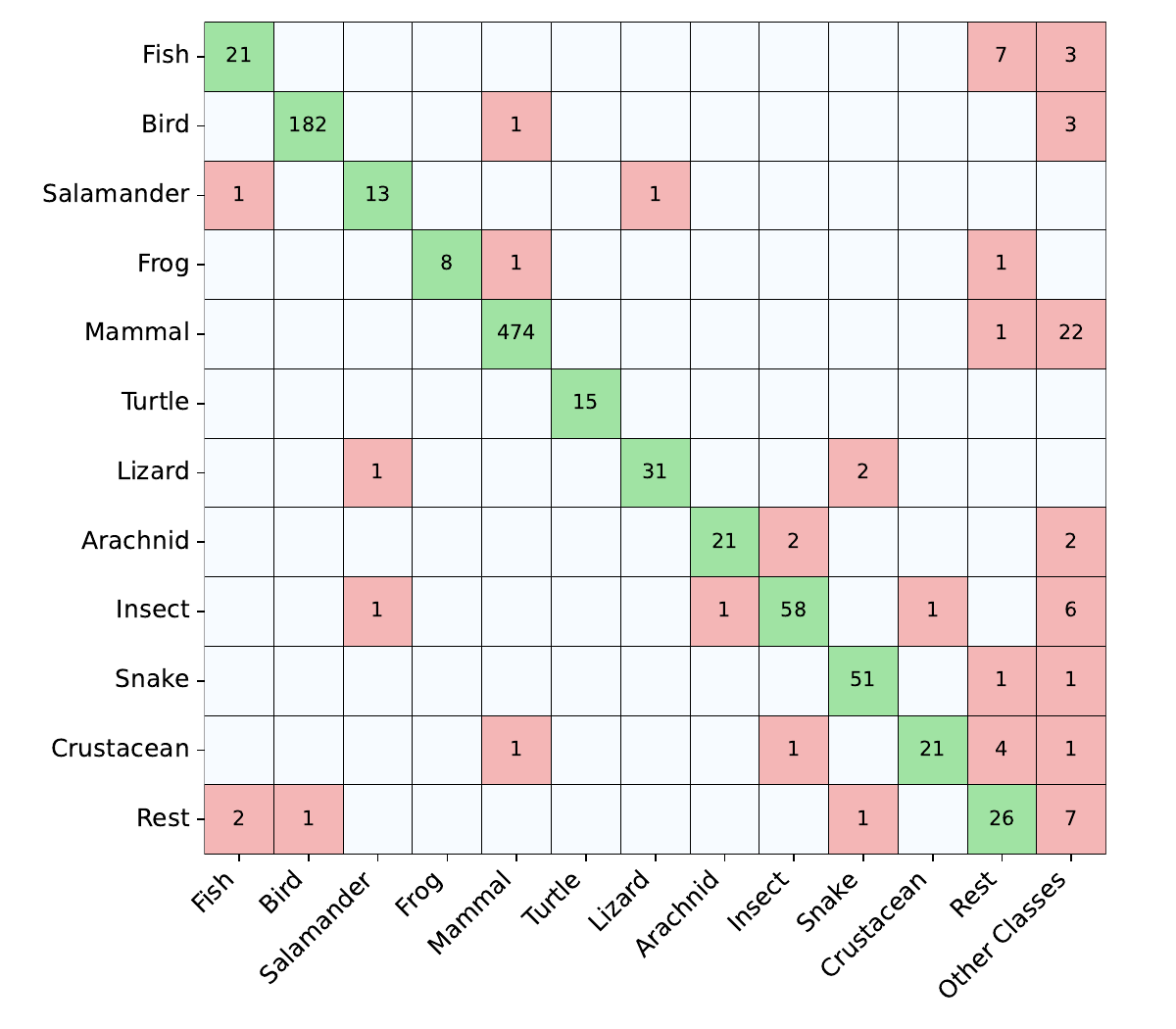}
        \caption{\label{fig:confusion_matrix}Confusion Matrix of HDC (Strategy 1) for the sub-classes under the synset class ``Animal''. The x-axis shows the predicted labels (including ``other classes'' outside of the synset class ``Animal''), and the y-axis shows the ground-truth labels.}
        
    \end{center} 
\end{figure}

In \autoref{fig:confusion_matrix}, we present a detailed confusion matrix of classes within the synset category “Animal.” 
Most misclassifications occur among biologically similar groups, such as Salamander-Lizard and Lizard-Snake, highlighting the classifier’s tendency to group closely related classes.

Overall, our results show that HDC provides a customizable trade-off between inference speed and accuracy, making it adaptable to varying application needs. 
Strategy 1 is particularly suitable for high-accuracy applications, while Strategy 2 is better suited to real-time scenarios that prioritize speed.

\section{Limitations \& Future Work}

While our method substantially improves inference time and maintains competitive accuracy, which is also interesting for robust and zero-shot open-set classification with dynamic class modifications (\textit{i.e.}, adding and removing class labels without additional training), several limitations warrant further investigation in future work.

Our ImageNet-1K experiments rely on a label tree derived from WordNet, but this prior knowledge is not strictly necessary. 
Alternative tree-construction methods could leverage LLMs to group similar labels into synsets, forming a hierarchy in a bottom-up manner, as shown in our CIFAR-100 experiments. 
Another approach could involve generating a candidate tree and refining it through greedy expansion.

Moreover, the efficiency gains provided by the hierarchical pruning strategy heavily depend on the depth and balance of the underlying label tree. Datasets with very shallow hierarchies or weak parent-child relationships may not benefit as significantly from our method.

\section{Conclusion}
In this work, we introduced the Hierarchical Diffusion Classifier (HDC), a novel approach for accelerating diffusion-based classification by utilizing hierarchical class pruning. Our results demonstrate that the HDC significantly reduces inference time, achieving up to a 60\% speedup over traditional diffusion classifiers while maintaining and, in some cases, even improving classification accuracy. This improvement is achieved by progressively narrowing down relevant class candidates, pruning out high-level categories early in the process, and focusing only on specific, contextually relevant subcategories.

Our experiments highlight HDC’s adaptability, showing that different pruning strategies (such as Top-k Pruning and Threshold Pruning) offer customizable trade-offs between inference speed and accuracy. 
This versatility makes HDC suitable for diverse applications, from high-accuracy image classification tasks to real-time scenarios where rapid inference is critical. Our work effectively expands the design space for DCs far beyond a straightforward combination and enables new research avenues.

\section{Societal Impact}
Despite the efficiency gains introduced by the Hierarchical Diffusion Classifier (HDC) for training-free image classification, diffusion classifiers remain computationally intensive, potentially leading to increased power consumption and a larger carbon footprint. 
This computational demand poses environmental concerns, especially when scaled to large datasets and real-time applications. 
Future research must prioritize developing even more energy-efficient architectures and optimization strategies to address these sustainability challenges, ensuring that the deployment of diffusion classifiers can be environmentally responsible.
The freedom to adopt various pruning strategies in HDC is a step toward into this direction. 

\section*{Acknowledgements}
This work was supported by the BMBF projects SustainML (Grant 101070408), Albatross (Grant 01IW24002), and by the Carl Zeiss Foundation through the Sustainable Embedded AI project (P2021-02-009).

{\small
\bibliographystyle{ieee_fullname}
\bibliography{egbib}
}

\end{document}